\documentclass{article}

\usepackage{arxiv}

\usepackage[utf8]{inputenc} 
\usepackage[T1]{fontenc}    
\usepackage[colorlinks=true,allcolors=blue]{hyperref}       
\usepackage{url}            
\usepackage{booktabs}       
\usepackage{amsfonts, amsmath, amssymb}       
\usepackage{nicefrac}       
\usepackage{microtype}      
\usepackage[round]{natbib}
\usepackage{faktor}
\usepackage{ntheorem}
\usepackage{graphicx}
\usepackage{enumitem}
\usepackage{color}
\usepackage{float}
\usepackage{algorithm} 
\usepackage[algo2e]{algorithm2e}

\renewcommand\cite{\citep}
\newtheorem{definition}{Definition}

\title{Notes on Coalgebras in Sylometry}

\author{Jo\"el A. Doat}

\begin{document}
\maketitle

\begin{abstract}
The syntactic behaviour of texts can highly vary depending on their contexts (e.g. author, genre, etc.). From the standpoint of stylometry, it can be helpful to objectively measure this behaviour. In this paper, we discuss how coalgebras are used to formalise the notion of behaviour by embedding syntactic features of a given text into probabilistic transition systems. By introducing the behavioural distance, we are then able to quantitatively measure differences between points in these systems and thus, comparing features of different texts. Furthermore, the behavioural distance of points can be approximated by a polynomial-time algorithm.
\end{abstract}


\section{Introduction}

In stylometry, syntactic features of a text can be used as an indicator for the author's style. In this research, a new method to objectively measure this behaviour is introduced. First, by embedding features into \textit{probabilistic transition systems} (PTSs) \cite{prob2}, one can obtain a statistical representation \cite{markov2}. These syntactic features include for example co-occurrence of words or word classes, parts-of-speech distribution, and grammatical structures \cite{stylo}. Then, the notion of behaviour in PTSs can be formalised by so-called \textit{probabilistic bisimulations} \cite{prob2}. These are used to contrast the probabilities of outgoing edges of two states in PTSs and return a value describing the alikeness of these states, i.e. a lower value indicates greater similarity. \\
When measuring the syntactical behaviour of a text with respect to a given feature, the idea is to associate a state of the corresponding PTS as a representant of the investigated text and compare it with the corresponding state in the system of another text. Thus, for every feature, we obtain a behavioural value for this text. \\
In order to calculate the value of probabilistic bisimulations, PTSs can be translated into \textit{coalgebras} \cite{jacobs} on metric spaces \footnote{spaces with distance measures} \cite{prob1}. Coalgebras can be seen as abstract dynamic transition systems consisting of a state space (metric space) and a transition function, describing one-step transitions at each state. The advantage of this representation is that PTSs admit a so-called \textit{terminal coalgebra} \cite{ultrametric} which intuitively describes a transition system realising all possible state behaviours while being a metric space itself again. In other words, when investigating texts, the terminal coalgebra can be seen as an abstract ontological space containing information (about the behaviour with respect to a given feature) of every possible text together with a distance measure. Thus, we can measure the so-called \textit{behavioural distances} \cite{breugel} between different behavioural realisations on the terminal coalgebra. Furthermore, by definition, for every state, there is then a unique map that associates it with its behavioural realisation. The map can be approximated by a polynomial-time algorithm and thus, the behavioural value of the investigated state. More specifically, this algorithm solves a \textit{transportation problem} \cite{linear} that can be reduced to a \textit{minimum cost network flow} \cite{network} and \textit{minimum cost circulation} \cite{circ} problem. \\
At last, this algorithm will be tested to obtain the first results. For our purpose, chosen texts will be analysed in contrast to the genre categories of the Brown corpus.

\section{Method}
\subsection{Probabilistic Transition System}
Probabilistic transition systems can be used as a method to represent statistical data with respect to the influence of its environment. The general idea is that two information states (e.g. lexical items in a text) can be related via a transitioning probability and a provoking action. Recall the formal definition.
\begin{definition} \cite{prob2}
    A probabilistic transition system (PTS) is a triple $(S,A,\pi)$ with a non-empty set of states $S$, a set of actions $A$ and a transition function $\pi:S \times A \times S \to [0,1]$ such that for all $s \in S$ and $a \in A$, $\sum_{s' \in S} \pi(s,a,s') \in [0,1]$.
\end{definition}

Note that PTSs without additional labels on the edges (unlabelled case) can be defined by a set of actions only containing an arbitrary element (i.e. $A:=\{\ast\}$). This allows to make no distinction between edges with respect to actions. For notational simplicity, we only consider the unlabelled case and denote PTSs as a tuple $(S,\pi)$. \\ 

Now, consider the following texts.
\begin{figure}[H] 
    \centering
	\includegraphics[width=0.6\textwidth]{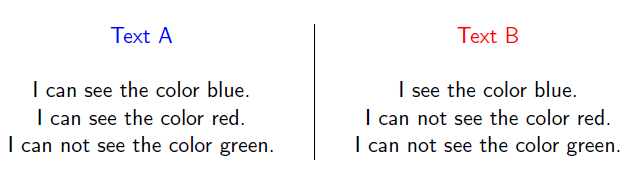}
\end{figure}

By labelling the edges of the corresponding word adjacency networks (WAN) with the transition probabilities, we obtain the following diagram. Note that the blue and and red arrows belong to the left and right text, respectively.  
\begin{figure}[H] 
    \centering
	\includegraphics[width=0.7\textwidth]{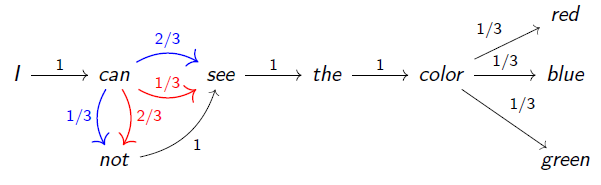}
	\caption{Probabilistic Transition System for Text A and B}
	\label{fig:pts1}
\end{figure}

As one can see in Fig.\ref{fig:pts1}, using probabilistic transition systems (PTSs), the lexical unit 'can' emits a different behaviour depending on the context which can be chosen in terms of linguistic features that can be embedded in PTSs. The resulting system represents a 'probabilistic reading flow' with respect to the given features. The question might arise by 'how much' the behaviour differs for a specific word. \\

For the unlabelled case (no additional actions), such models are already in use for stylometric analysis, e.g. sequence of characters \cite{markov1}, sequence of words \cite{markov2} and sequence of function words \cite{markov3}. Usually, this is done by averaging values of a transition matrix given by a text and comparing different texts with these results. Since these systems are used to represent the syntactic style of a text, the corresponding behaviour can be seen as a core property of this style representation. \\

\subsection{Probabilistic Bisimulation}

In this context, the qualitative equivalence notion to check if emitted behaviour between states of a PTS coincide is called \textit{probabilistic bismulation}.

\begin{definition} \cite{prob2}
    For a set of states $S$ and transition function $\pi$ define a probabilistic transtion system $(S,\pi)$.   An equivalence relation $\mathcal{R}$ on the set of states $S$ is a
	probabilistic bisimulation if $s\mathcal{R}s'$ implies $\sum_{\bar{s} \in E} \pi(s,\bar{s}) = \sum_{\bar{s} \in E} \pi(s',\bar{s})$ for all $\mathcal{R}$-equivalence classes $E$. States
	$s$ and $s'$ are probabilistic bisimilar if $s\mathcal{R}s'$ for some probabilistic bisimulation $\mathcal{R}$.
\end{definition}

Such a bisimulation is intended to be a witness for equivalent behaviour between states, i.e. two states are equivalent if they have the same transitioning probability to every equivalence class of the bisimulation. \\

Consider the follwoing PTS.
\begin{figure}[H]
    \centering
	\includegraphics[width=0.5\textwidth]{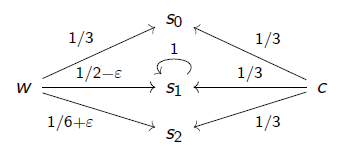}
	\caption{Probabilistic Transition System t compare behaviour of $w$ and $c$}
	\label{fig:pts2}
\end{figure}
	
In Fig.\ref{fig:pts2}, the states $w$ and $c$ emit the same  iff $\varepsilon=1/6$. Thus, in this simple case, we would like to define the difference between $w$ and $c$ as $1/6$ if $\varepsilon=0$. Motivated by this definition, \cite{breugel} use a pseudometric\footnote{A pseudometric is like a usual metric with the difference that two distinct elements can have distance 0.} on states of PTSs as a method to witness a quantitative version of this behavioural equivalence, i.e. a smaller distance signifies greater similarity. The idea is to represent these PTSs as \textit{coalgebras} \cite{jacobs} on metric spaces\footnote{More precisely, PTSs can be seen as coalgebras on the category of $1$-bounded complete ultrametric spaces and non-expansive functions \cite{prob1}.}. The advantage of this representation is that PTSs admit a so-called \textit{terminal coalgebra}\footnote{See Rutten and Turi's ultrametric terminal coalgebra theorem \cite{ultrametric}} which intuitively describes a transition system realising all possible state behaviours, while being a metric space itself again. By definition of terminal coalgebras, for every state there is then a unique map that associates it with its behavioural realisation. The distance between such realisations is called \textit{behavioural distance}.

\subsection{Coalgebra}\label{sec:coalgebra}

In this subsection, I introduce coalgebras and the behavioural distances in a rather intuitive way to focus on how this idea interacts with stylometry instead of introducing all necessary terms from category theory. 

\begin{definition}\cite{jacobs}
    For an (set) endofunctor $T$, object (set) $C$, and morphism $\gamma:C\to TC$ the tuple $(C,\gamma)$ is called a \textit{$T$-coalgebra}.
\end{definition}

Coalgebras can be seen as abstract state-based transition systems. It is a more general notion than PTSs because it does not restrict to a set of states with transitioning probabilities between them. Such a $T$-coalgebra $(C,\gamma)$ consists of three parameters that describe it. 
\begin{enumerate}
    \item[\textbullet] The object (set) $C$ is called the \textit{carrier} of a coalgebra and denotes the states of the transition system. In the case of PTSs, $C$ is a set of states.
    \item[\textbullet] The \textit{type} $T$ describes all observable behaviours after one step. Applying this endofunctor transforms the object of a type (e.g. set) into an object of the same type. For PTSs, $T$ is the so-called distribution functor which maps a set of states $C$ to a set $TC$ of all possible probabilistic distributions on the same states. $TC$ is also referred to as \textit{structured successor}.
    \item[\textbullet] $\gamma$ denotes the actual transition function of the system, i.e. mapping one element of the carrier to the behaviour it is emitting. In PTSs, this is a function mapping a state to its corresponding probabilistic distribution. 
\end{enumerate}

Since the set $TC$ contains all possible observable behaviours of states in $C$ after one step, we can apply $T$ again and obtain with $T^2C$ the observable behaviours after two steps. This process can be continued iteratively to describe the behaviour after arbitrarily many steps. \\
As mentioned in the last subsection, the existence of a terminal coalgebra guarantees that the information about the behaviour of $(C,\gamma)$ collapses through the iteration process into a single point inside the terminal coalgebra itself. In other words, the terminal coalgebra realises every possible $n$-step behaviour with respect to $T$ and thus, there is one state that emits the same behaviour as a state in the coalgebra we started with. Therefore, states in the terminal coalgebra can be seen as behavioural realisations of transitions systems.

\begin{figure}[H]
    \centering
	\includegraphics[width=0.6\textwidth]{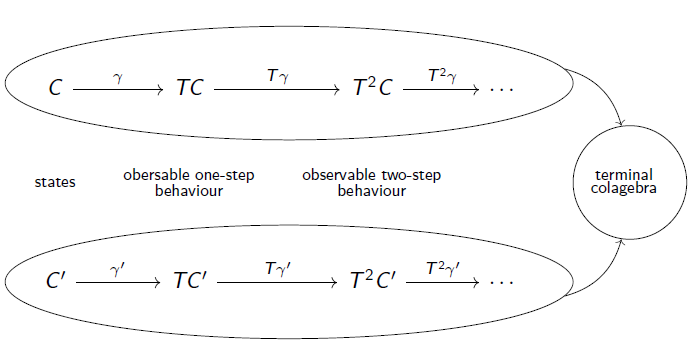}
	\caption{Observable behaviour collapses to one point in the terminal coalgebra}
	\label{fig:terminal}
\end{figure}

As PTSs are special instances of such abstract transition systems, we can embed the statistical representation of texts (see Fig.\ref{fig:pts1}) into a coalgebra. In Fig.\ref{fig:terminal}, ($C$,$\gamma$) and $(C',\gamma')$ constitute now two texts as coalgebras. For a chosen starting point in the PTSs the observable behaviours of these texts with respect to the representation collapse to point on the terminal coalgebra. To measure the behavioural distance which correspond to a stylistic distance in our context, we need a metric on the terminal coalgebra. \cite{breugel} showed that PTSs can be equivalently represented as transition systems on pseudometric spaces such that also the distance of probabilistic distributions in the structured successors can be measured. The advantage of this is that terminal coalgebra also becomes a pseudeometric space and thus, behavioural/stylistic distances between texts exist. In other words, when investigating texts, this terminal coalgebra is a naive approach to build an ontology containing behavioural/stylistic information (with respect to a given syntactic features) of every possible text together with a distance measure.   

\subsection{Algorithm}
On the basis of the idea discussed in Section \ref{sec:coalgebra}, \cite{breugel} extracted a polynomial algorithm that approximates the value of behavioural distances. The algorithm solves a linear programming problem.

\begin{figure}[H]
    \centering
    \includegraphics[width=0.4\textwidth]{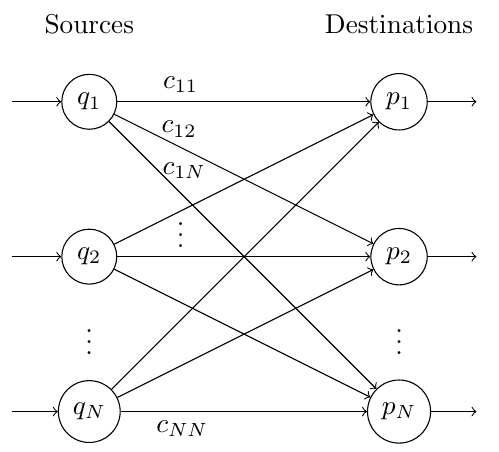}
    \caption{Network flow model of the transportation problem}
    \label{fig:transportation}
\end{figure}

In a PTS with $N$-many states, the outgoing probabilities of two states $p_1,\dots,p_N$ and $q_1,\dots,q_N$, respectively, are compared by solving a transportation problem. In Fig.\ref{fig:transportation}, one can see the corresponding network flow model. The idea is to find the minimal transportation costs $c_{i,j},i,j=1,\dots,N$ from the source $q_i$ to the destination $p_j$ where the outgoing probabilities label the states with their supply (source) and demand (destination). To encapsulate also more than one transition step and thus, increase the accuracy of the result, this process can be iterated arbitrarily many times\footnote{One can imagine the iteration process in each step as an additional comparison along the observable $n$-step behaviour as shown in Fig.\ref{fig:terminal}.}. Furthermore, thanks to \textbf{Proposition 12} in \cite{breugel}, the accuracy $\alpha$ can be chosen freely since for some constant $c\in (0,1)$\footnote{The value of this constant is prescribed by the proposition.} the difference between the value of the behavioural distance at step $\lceil log_c(\alpha / 2) \rceil$ and the actual distance $d$ is smaller or equal to $\alpha$. Note that the calculated result is then $c\cdot d$. \\

\SetKwFor{RepTimes}{repeat}{times}{end}
\begin{figure}[H]
    \begin{algorithm}[H]
    \DontPrintSemicolon
    \tcp*[h]{Step 1: Initialisation} \\
    \ForEach{$1 \leq k,l \leq N$}{
        $d_{kl} \gets 0$
    } 
    $c_{00} \gets 0$ \\
    \ForEach{$1 \leq i,j \leq N$}{
        $c_{i0} \gets 1$ \\
        $c_{0j} \gets 1$
    }
    \tcp*[h]{Step 2: Main loop} \\
    \RepTimes{$\lceil log_c(\alpha / 2) \rceil$}{ 
        \ForEach{$1 \leq i,j \leq N$}{
            $c_{ij} \gets c\cdot d_{ij}$
        }
        \ForEach{$1 \leq k,l \leq N$}{
            \tcc{\small min\_value defined below}
            $d_{kl} \gets \text{min\_value}(k,l)$
            
        }
    }
    
    \end{algorithm}
    \caption{Pseudocode of the algorithm}
    \label{fig:algorithm}
\end{figure}

Fig.\ref{fig:algorithm} describes the algorithm in pseudocode. For a PTS $(S,\pi)$ with states $S=\{s_1,\dots,s_N\}$, the operation min\_value$(k,l)$ is defined in each iteration step by the minimal value of 
\begin{align*}
    \sum_{i,j=0}^{N}c_{ij}\lambda_{ij}
\end{align*}
with constraints 
\begin{align*}
    &\sum_{i,j=0}^{N}\lambda_{ij}=\pi(s_k,s_j), & 0\leq j \leq N,  \\
    &\sum_{i,j=0}^{N}\lambda_{ij}=\pi(s_l,s_i), & 0\leq i \leq N, \\
    &0\leq \lambda_{ij},\, 0\leq i,j\leq N,
\end{align*}
where
\begin{align*}\label{eq:sub}
    \pi(s_i,s_0):= 1-\sum_{j=1}^{N} \pi(s_i,s_j).
\end{align*}
The resulting matrix $d$ is the \textit{distance matrix}, i.e. the distance between states $s_k$ and $s_l$ is the value $d_{kl}$.

\section{Application}
\subsection{Comparison of two texts}
To be able to use the algorithm on two different texts, we need to combine the corresponding PTSs into the same system. As we saw in Fig.\ref{fig:pts2}, the idea of behavioural distance is to compare transition probabilities to the same equivalence classes of states. This means that the easiest way to compare texts is to connect them at some chosen end state. In other words, all transition walks of both PTSs can be extended to a walk whose final node corresponds to the designated end state and the comparison of states materialise in terms of their relation to it. \\
Secondly, we need states that represent the whole text in each PTS such that they can be juxtaposed to measure a stylistic distance. 

\begin{figure}[H]
    \centering
	\includegraphics[width=0.7\textwidth]{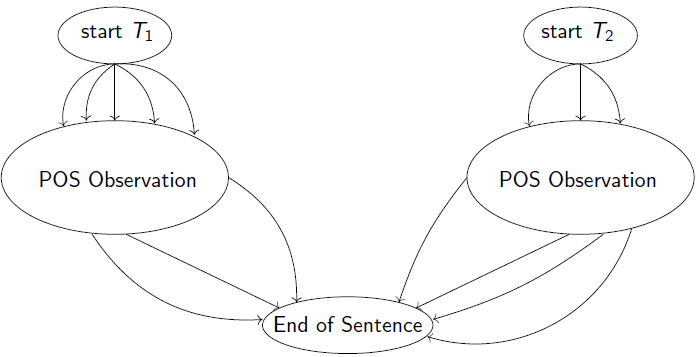}
	\caption{How to compare two texts}
	\label{fig:idea}
\end{figure}
Fig.\ref{fig:idea} demonstrates an example on how to apply this method for the purpose of stylometry. Given texts $T_1$ and $T_2$, one could measure the difference between the distribution of Parts-of-Speech (POS), i.e. a PTS with POS as states and the transitioning probabilities given by their co-occurrences in the text. In order to combine  both PTSs into one, the punctuation marks at the end of the sentences are represented by a single state which coincides for the two texts. Each state representing a POS has then a transition to this end state labelled by the probability that a sentence ends after given POS. This allows a clear comparison by using the end state as the only state in the equivalence class accessible by both systems. Next, we add a starting state in each PTS that has transitions to each state representing a POS that occur at the beginning of a sentence and label these transitions with the probability that a sentence starts with the given POS. The two starting states can then be compared in terms of the behavioural distance. The resulting value represents the syntactical difference for the POS distributions. \\
Further linguistic features/co-occurrence phenomenons that can be represented as such a transition system include: lexical items, letters, collocations of specific POS (e.g. nouns after modal verbs), function words, and grammar trees of sentences. In the next section, I used multiple features to obtain a more precise result. More specifically, after the same principle as in Fig.\ref{fig:idea}, I also used the letter distribution (states represent letters in the observation instead of POS) and the grammar trees (states represent grammatical constituents in the hierarchy). The corresponding starting states connect to the first letters in sentences and the first separation into phrasal constituents in the grammatical hierarchy of sentences, respectively. The end state is given by punctuation marks and the lowest level in the grammar trees\footnote{For both PTSs that represent the statistical relation of grammar trees in the texts, the last level in the hierarchy corresponds to the actual lexical items that are irrelevant for the grammatical observation. Thus, all levels consisting of lexical items collapse to the end state of the combined system.}, respectively.       \\

\subsection{Results}
To demonstrate the idea, I applied the algorithm in the context of genre classification. For this purpose, we take 4 texts and compare each of them with texts that represent genres as discussed in the last subsection (see Fig.\ref{fig:idea}). The genres are \textit{news}, \textit{religion}, \textit{government}, and \textit{belles-lettres} from the Brown corpus. The texts are the coconut news (Reuters corpus), the bible (Gutenberg corpus), Donald Trump's inauguration speech (Inaugural Address corpus), and Lewis Carroll's Alice's Adventures in Wonderland (Gutenberg Corpus). For all comparisons we investigate the syntactic features Parts-of-Speech, grammar trees, and letters. The three resulting values can be seen as a vector whose distance to $0$ can be calculated with the euclidean distance in order to compare it with other categories. The results can be found in Fig.\ref{fig:results1}-\ref{fig:results4}.

\begin{figure}[H]
    \centering
    \begin{tabular}{|p{1.3cm}|p{1cm}|p{1cm}|p{1cm}|p{1cm}|}
         \hline
         & news & religion & gov. & belles   \\ 
         \hline
         POS & 0.2778 & 0.3899 & 0.3689 & 0.41  \\
         \hline
         Grammar & 0.322  & 0.4111 & 0.3556 & 0.4444 \\
         \hline 
         Letters & 0.083 & 0.0967 & 0.0877 & 0.1222 \\
         \hline 
         \hline
         Euclid & 0.4333 & 0.5748 & 0.5199 & 0.6166 \\
         \hline
    \end{tabular}
    \caption{Results for coconut news}
    \label{fig:results1}
\end{figure}
\begin{figure}[H]
    \centering
    \begin{tabular}{|p{1.3cm}|p{1cm}|p{1cm}|p{1cm}|p{1cm}|}
         \hline
         & news & religion & gov. & belles   \\ 
         \hline
         POS & 0.3661 & 0.2795 & 0.2633 & 0.314  \\
         \hline
         Grammar & 0.4333  & 0.2919 & 0.3201 & 0.4005 \\
         \hline 
         Letters & 0.1007 & 0.0501 & 0.0199 & 0.0885 \\
         \hline 
         \hline
         Euclid & 0.5761 & 0.4072 & 0.415 & 0.5166 \\
         \hline
    \end{tabular}
    \caption{Results for the bible}
    \label{fig:results2}
\end{figure}
\begin{figure}[H]
    \centering
    \begin{tabular}{|p{1.3cm}|p{1cm}|p{1cm}|p{1cm}|p{1cm}|}
         \hline
         & news & religion & gov. & belles   \\ 
         \hline
         POS & 0.2519 & 0.2238 & 0.2115 & 0.2903  \\
         \hline
         Grammar & 0.2581  & 0.273 & 0.2667 & 0.3856 \\
         \hline 
         Letters & 0.0447 & 0.0329 & 0.0454 & 0.0548 \\
         \hline  
         \hline
         Euclid & 0.359 & 0.3545 & 0.3434 & 0.4858 \\
         \hline
    \end{tabular}
    \caption{Results for Donald Trump's inauguration speech}
    \label{fig:results3}
\end{figure}
\begin{figure}[H]
    \centering
    \begin{tabular}{|p{1.3cm}|p{1cm}|p{1cm}|p{1cm}|p{1cm}|}
         \hline
         & news & religion & gov. & belles   \\ 
         \hline
         POS & 0.2868 & 0.2175 & 0.2284 & 0.2273    \\
         \hline
         Grammar & 0.3781  & 0.3754 & 0.3502 & 0.3389 \\
         \hline 
         Letters & 0.0732 & 0.0459 & 0.0177 & 0.0638 \\
         \hline 
         \hline
         Euclid & 0.4802 & 0.4363 & 0.4185 & 0.413 \\
         \hline
    \end{tabular}
    \caption{Results for Alice's Adventures in Wonderland}
    \label{fig:results4}
\end{figure}

As one can see in these results, the distance between a text and its correct category is the lowest. Thus, for this small experiment the classification task worked. In the context of this demonstration, it is clear that the method of behavioural distance gives an interesting research direction for stylometry. Nevertheless, further experiments are still required to check if this method is reliable since the difference between some distances is rather small (see euclidean distances to government and belles-lettres in Fig.\ref{fig:results4}). 

\section{Conclusion}
In this paper, I introduced a new tool for objective measurement of text behaviour. By embedding syntactic features of texts into PTSs and using the method of \cite{breugel}, we have shown how to calculate the distance of behaviour realisations. Furthermore, a polynomial-time algorithm exists to approximate this distance and according to the chosen syntactic feature, it delivers a vector-like representation of the investigated text. In a small experiment we have seen a small task of genre classification which has been solved correctly with this method. \\
There are several aspects that can be improved and are subject to future work. So far, I only used the features regarding Parts-of-Speech, grammar trees, and letters. As already mentioned, further features that can be represented as PTSs include lexical items, collocations of specific POS (e.g. nouns after modal verbs), and function words. \\
In Fig.\ref{fig:idea}, we saw one idea of how two PTSs can be combined. Since this combination is variable, it remains to be investigated if there are better ways to achieve this. Another direction one could possibly go is to embed a PTS into a canonical construction instead, i.e. extending every PTS uniformly into a bigger system such that every text obtain a behavioural distance value with respect to the same construction without a second text. \\
Two properties of PTSs we did not make use of in this paper are actions and subprobabilities. 
\begin{figure}[H]
    \centering
    \includegraphics[width=0.4\textwidth]{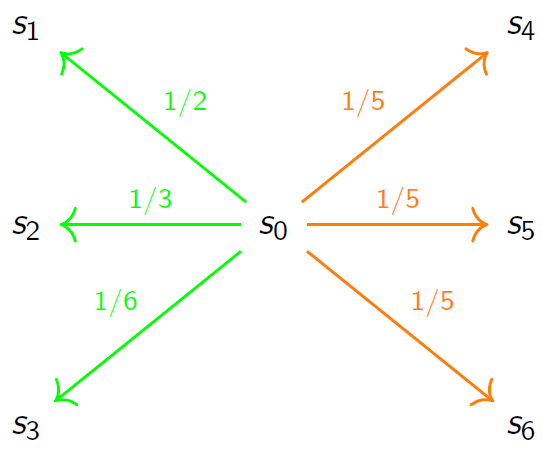}
    \caption{Labels and subprobabilities}
    \label{fig:label}
\end{figure}
Fig.\ref{fig:label} demonstrates a simple example of such a case. The colors green and orange denote two different actions on $s_0$ that have distinct probability distributions. According to \cite{breugel} the algorithm can also be extended for actions. Furthermore, the sum of all probabilities along the orange action is smaller than $1$. This can be interpreted as an additional probability that the system refuses to interact with the environment at $s_0$. By adding states representing punctuation marks like comma, colon, and semicolon, one can include subprobability distributions at these states to denote an interruption in the reading flow. \\   
An open question not covered in this paper is the possible use of this method in other contexts like for example pragmatics or register analysis. A state representing a concrete lexical item can be introduced and the corresponding PTS describes a co-occurrence phenomenon (e.g. parts-of-speech occurring after the modal verb "can"). This can be useful measure the behavioural distance of this item in different contexts.

\newpage
\bibliographystyle{plainnat}  
\bibliography{main}

\end{document}